\documentclass[lettersize,journal]{IEEEtran}
\usepackage{amsmath,amsfonts}
\usepackage{amsthm}
\theoremstyle{definition}
\newtheorem{definition}{Definition}
\usepackage{algorithmic}
\usepackage{algorithm}
\usepackage{array}
\usepackage[caption=false,font=normalsize,labelfont=sf,textfont=sf]{subfig}
\usepackage{textcomp}
\usepackage{stfloats}
\usepackage{url}
\usepackage{verbatim}
\usepackage{graphicx}
\usepackage{cite}
\usepackage{cuted}
\usepackage{lipsum}
\DeclareMathOperator*{\argmax}{arg\,max}
\DeclareMathOperator*{\argmin}{arg\,min}
\usepackage[table]{xcolor}

\newcommand*{\argminl}{\argmin\limits}

\newcommand\Tstrut{\rule{0pt}{2.6ex}}         
\newcommand\Bstrut{\rule[-0.9ex]{0pt}{0pt}}   
\begin{document}

\title{High-Order Multilinear Discriminant Analysis via Order-\textit{n} Tensor Eigendecomposition}

\author{Cagri Ozdemir~\IEEEmembership{Student Member,~IEEE,} Randy C. Hoover, Kyle Caudle, and Karen Braman
\thanks{The current research was supported in part by the Department of the Navy, Naval Engineering Education Consortium under Grant No. (N00174-19-1-0014) and the National Science Foundation under Grant No. (2007367).}
\thanks{Cagri Ozdemir and Randy C. Hoover are with the Department of Computer Science and Engineering, South Dakota Mines, Rapid City, SD 577701 USA (e-mail: cagri.ozdemir@mines.sdsmt.edu; randy.hoover@sdsmt.edu).}
\thanks{Kyle Caudle and Karen Braman are with the Department of Mathematics, South Dakota Mines, Rapid City, SD 57701 USA (e-mail: kyle.caudle@sdsmt.edu;karen.braman@sdsmt.edu).}}



\maketitle

\begin{abstract}
Higher-order data with high dimensionality is of immense importance in many areas of machine learning, computer vision, and video analytics. Multidimensional arrays (commonly referred to as tensors) are used for arranging higher-order data structures while keeping the natural representation of the data samples. In the past decade, great efforts have been made to extend the classic linear discriminant analysis for higher-order data classification generally referred to as multilinear discriminant analysis (MDA). Most of the existing approaches are based on the Tucker decomposition and \textit{n}-mode tensor-matrix products. The current paper presents a new approach to tensor-based multilinear discriminant analysis referred to as High-Order Multilinear Discriminant Analysis (HOMLDA). This approach is based upon the tensor decomposition where an order-\textit{n} tensor can be written as a product of order-\textit{n} tensors and has a natural extension to traditional linear discriminant analysis (LDA). Furthermore, the resulting framework, HOMLDA, might produce a within-class scatter tensor that is close to singular. Thus, computing the inverse inaccurately may distort the discriminant analysis. To address this problem, an improved method referred to as Robust High-Order Multilinear Discriminant Analysis (RHOMLDA) is introduced. Experimental results on multiple data sets illustrate that our proposed approach provides improved classification performance with respect to the current Tucker decomposition-based supervised learning methods. 
\end{abstract}

\begin{IEEEkeywords}
High-Order Multilinear Discriminant Analysis, Robust High-Order Multilinear Discriminant Analysis, tensor eigendecomposition, multilinear subspace learning.
\end{IEEEkeywords}

\section{Introduction}
\label{introduction}
\IEEEPARstart{H}{igher} order data appears in many real-world applications such as image classification, video analytics, and pattern recognition. These types of multidimensional data can be represented as multidimensional arrays of numbers, commonly referred to as tensors~\cite{de1998matrix,kolda2009tensor,nagy2006kronecker}. The dimensions of the tensor are generally referred to as ways or modes. The number of modes determines the order of a tensor. For example, a color image is a third-order tensor (row pixels, column pixels, and color channels) and a video that consists of color images is a fourth-order tensor (row pixels, column pixels, color channels, and time).

Linear discriminant analysis (LDA) is one of the most popular shallow learning algorithms for feature extraction and has been widely used for subspace learning and dimensionality reduction~\cite{hart2000pattern,belhumeur1997eigenfaces,fisher1936use,yan2014multitask}. When dealing with image data, each observation must first be converted to a vector and stacked temporally to form a 2-mode tensor (matrix in this case).  However, converting higher-order tensors into vectors brings about two major issues: 1) Vectorizing higher-order data samples eliminates the spatial correlations within each sample and 2) in general the number of samples is relatively small compared to the feature vector dimension. That causes a small sample size problem and may cause the scatter matrices of LDA to become singular~\cite{fukunaga2013introduction,sharma2015linear}. In order to solve these problems, new methods have been employed that rely on higher-order data structures, leaving each sample in its natural tensor form. Discriminant analysis with tensor representation (DATER)~\cite{yan2005discriminant} and general tensor discriminant analysis (GTDA)~\cite{tao2007general} provide iterative procedures to maximize the scatter ratio criterion. However, DATER does not converge over iteration. Although the iterative approximation method of GTDA converges, it converges to a local maximum. Thus, two effective algorithms, direct general discriminant analysis (DGTDA) and constrained multilinear discriminant analysis (CMDA) are proposed as outlined in~\cite{li2014multilinear}. DGTDA learns tensor subspaces by obtaining the global maximum scatter ratio without iteration, whereas CMDA guaranties the convergence over iterations. All these supervised learning methods relies on Tucker structure and the \textit{n}-mode product~\cite{de2000multilinear,kolda2009tensor}.

Tensor singular value decomposition (t-SVD)~\cite{kilmer2008third,braman2010third,kilmer2011factorization} provides a tensor decomposition for third-order tensors. Fundamental to t-SVD is 
the defined multiplication operator on third-order tensors (t-product) based upon Fourier theory and an algebra of circulants~\cite{karner2003spectral,gleich2013power}. Pose estimation and image classification applications were proposed in which principal component analysis (PCA) has been extended to third-order tensors to learn multilinear mappings from a third-order tensor input~\cite{hoover2011pose,kilmer2013third,hao2013facial,ozdemir20212dtpca}. The extension of the t-SVD to order-\textit{n} tensors was formulated and this extended structure was used in image deblurring and video facial recognition applications~\cite{martin2013order}. As the t-product is a convolution-like operation which can be implemented using the Fast Fourier Transform (FFT), variations on the classical t-product have been investigated in~\cite{kernfeld2015tensor} where it is shown that a family tensor-tensor products can be defined directly in a transform domain for an arbitrary invertible linear transform. Most recently, fast algorithms for the t-product and t-SVD have been developed~\cite{tarzanagh2018fast,ozdemir2021fast}. Finally, the t-product (and the many variations therein) have been applied to tensor completion~\cite{zhang2016exact,zhou2017tensor,song2020robust,qin2022low} and multilinear image processing~\cite{soltani2016tensor,zhang2018nonlocal,tarzanagh2018fast}. 

Although the t-SVD and t-product have been applied to various unsupervised learning tasks, supervised learning tasks have not been fully explored. In this paper, we introduce a new framework reffered to as High-Order Multilinear Discriminant Analysis (HOMLDA) for order-\textit{n} tensors based on tensor-tensor decomposition as opposed to utilizing a Tucker representation. Our proposed approach builds upon our prior multilinear discriminant analysis (MLDA) approach defined for third-order tensors~\cite{hoover2018multilinear}. Moreover, the within-class scatter tensor computed for HOMLDA may be close to singular and cause poor classification performance. To overcome this issue, a novel Robust High-Dimensional Multilinear Discriminant Analysis (RHOMLDA) is proposed. 

The rest of the paper is organized as follows. In Section~\ref{mathpre}, we discuss the mathematical foundations of the tensor operators and the tensor-tensor eigendecomposition. In Section~\ref{sec:MLDA}, we review the classic LDA method and propose our new frameworks for HOMLDA and RHOMLDA. In Section~\ref{experimental}, we compare the proposed methods with the state-of-art Tucker structure based discriminant analysis methods for classification applications, and Section~\ref{sec:conc} concludes the paper.

\section{Mathematical Preliminaries}
\label{mathpre}
As outlined in~\cite{kilmer2008third,braman2010third,kilmer2011factorization}, the tensor operators have been introduced for tensors of order three based on the discrete Fourier transform (DFT). The DFT based tensor operators have been extended to order-\textit{n} tensors in~\cite{martin2013order}. Furthermore, a family of tensor-tensor products for third-order tensors has been formulated by appealing to the transform domain~\cite{kernfeld2015tensor}. In this section, we extend these works and introduce a family of tensor operators for order-\textit{n} tensors defined directly in the transform domain via any discrete invertible linear transform. Finally, for completeness, we provide the tensor-tensor eigendecomposition for order-\textit{n} tensors built on the defined tensor operators.

\subsection{Notation}
\label{subsec:notation}
First, we review the basic definitions from
\cite{kernfeld2015tensor} and~\cite{kolda2009tensor} and introduce some basic notation. Capital script notation is used to refer to tensors such as $\mathcal{A}$, $\mathcal{B}$; a frontal slice of an order-\textit{n} tensor will be in capital script with a subscript indexing such as $\mathcal{A}_{(i_1...i_n)}$, $\mathcal{B}_{(i_1...i_n)}$. For example, let $\mathcal{A} \in \mathbb{R}^{m_1 \times m_2 \times m_3 \times m_4}$ be a fourth-order tensor. In terms of  {\sc Matlab}
indexing notation, $\mathcal{A}_{(i_3i_4)}=\mathcal{A}(:,:,i_3,i_4) \in \mathbb{R}^{m_1 \times m_2}$ is the frontal slice corresponding to $i_3^{\text{th}}$ and $i_4^{\text{th}}$ index of mode-3 and mode-4 respectively.
A lateral slice of  a tensor $\mathcal{A} \in \mathbb{R}^{m_1 \times m_2 \times m_3 \times m_4}$ is denoted as $\Vec{\mathcal{A}}_{(i_2)} = \mathcal{A}(:,i_2,:,...,:) \in \mathbb{R}^{m_1 \times 1 \times m_3 \times m_4}$.
We use capital non-script notation to denote matrices (2-mode tensors) such as $\mathbf{A}$, $\mathbf{B}$.
We use the notation $\mathbf{a}_{i_1..i_n}^{(k)}$ to denote the mode-$k$ fiber corresponding to $i_1^{\text{th}}$,..., $i_n^{\text{th}}$ index of an order-\textit{n} tensor. For example, let $\mathcal{A} \in \mathbb{R}^{m_1 \times m_2 \times m_3 \times m_4}$ be a fourth-order tensor, then $\mathbf{a}_{i_2i_3i_4}^{(1)}= \mathcal{A}(:,i_2,i_3,i_4)$ $ \in \mathbb{R}^{m_1}$ is the mode-$1$ fiber corresponding to $i_2^{\text{th}}$, $i_3^{\text{th}}$, and  $i_4^{\text{th}}$ index of  mode-$2$, mode-$3$, and mode-$4$ respectively. Similarly, mode-$k$ fibers can be obtained by fixing mode-$k$ dimension. For example,  $\mathbf{a}_{i_1i_3i_4}^{(2)}= \mathcal{A}(i_1,:,i_3,i_4)$ $ \in \mathbb{R}^{m_2}$, $\mathbf{a}_{i_1i_2i_4}^{(3)}= \mathcal{A}(i_1,i_2,:,i_4)$ $ \in \mathbb{R}^{m_3}$, and $\mathbf{a}_{i_1i_2i_3}^{(4)}= \mathcal{A}(i_1,i_2,i_3,:)$ $ \in \mathbb{R}^{m_4}$ denote mode-$2$, mode-$3$, and mode-$4$ fibers of the tensor $\mathcal{A}$ respectively.
\begin{definition}
\label{def:modekfold/unfold}
 Let $\mathcal{A} \in \mathbb{R}^{m_1\times m_2\times....\times m_n}$ be an order-\textit{n} tensor. Then $\texttt{unfold}^{(\texttt{k})}(\mathcal{A})$ maps the tensor $\mathcal{A}$ into a $m_k\times(m1...m_{k-1}m_{k+1}...m_n)$ matrix by stacking all the mode-$k$ fibers of the tensor $\mathcal{A}$ as the columns of the resultant matrix. The operation that takes $\texttt{unfold}^{(\texttt{k})}(\mathcal{A})$ back to tensor form is the $\texttt{fold}^{(\texttt{k})}$ command:
 \[
 \mathcal{A} = \texttt{fold}^{(\texttt{k})}\big(\texttt{unfold}^{(\texttt{k})}(\mathcal{A})\big).
 \]
\end{definition}

We briefly illustrate mode-$k$ unfolding with an example. Let $\mathcal{A} \in \mathbb{R}^{2 \times 2 \times 2 \times 2}$ be a mode-4 tensor.

 Mode-1 unfolding $\texttt{unfold}^{(\texttt{1})}(\mathcal{A}) \in \mathbb{R}^{2 \times 8}$ is:
\begin{align*}
&\texttt{unfold}^{(\texttt{1})}(\mathcal{A})\\
&=
\begin{pmatrix}
\mathbf{a}_{111}^{(1)}&\mathbf{a}_{211}^{(1)}& \mathbf{a}_{121}^{(1)}&\mathbf{a}_{221}^{(1)}& \mathbf{a}_{112}^{(1)}&\mathbf{a}_{212}^{(1)}& \mathbf{a}_{122}^{(1)}&\mathbf{a}_{222}^{(1)}
\end{pmatrix}.
\end{align*}

 Mode-2 unfolding $\texttt{unfold}^{(\texttt{2})}(\mathcal{A}) \in \mathbb{R}^{2 \times 8}$ is:
\begin{align*}
&\texttt{unfold}^{(\texttt{2})}(\mathcal{A})\\
&=
\begin{pmatrix}
\mathbf{a}_{111}^{(2)}&\mathbf{a}_{211}^{(2)}& \mathbf{a}_{121}^{(2)}&\mathbf{a}_{221}^{(2)}& \mathbf{a}_{112}^{(2)}&\mathbf{a}_{212}^{(2)}& \mathbf{a}_{122}^{(2)}&\mathbf{a}_{222}^{(2)}
\end{pmatrix}.
\end{align*}

Mode-3 unfolding $\texttt{unfold}^{(\texttt{3})}(\mathcal{A}) \in \mathbb{R}^{2 \times 8}$ is:
\begin{align*}
&\texttt{unfold}^{(\texttt{3})}(\mathcal{A})\\
&=
\begin{pmatrix}
\mathbf{a}_{111}^{(3)}&\mathbf{a}_{211}^{(3)}& \mathbf{a}_{121}^{(3)}&\mathbf{a}_{221}^{(3)}& \mathbf{a}_{112}^{(3)}&\mathbf{a}_{212}^{(3)}& \mathbf{a}_{122}^{(3)}&\mathbf{a}_{222}^{(3)}
\end{pmatrix}.
\end{align*}

Mode-4 unfolding $\texttt{unfold}^{(\texttt{4})}(\mathcal{A}) \in \mathbb{R}^{2 \times 8}$ is:
\begin{align*}
&\texttt{unfold}^{(\texttt{4})}(\mathcal{A})\\
&=
\begin{pmatrix}
\mathbf{a}_{111}^{(4)}&\mathbf{a}_{211}^{(4)}& \mathbf{a}_{121}^{(4)}&\mathbf{a}_{221}^{(4)}& \mathbf{a}_{112}^{(4)}&\mathbf{a}_{212}^{(4)}& \mathbf{a}_{122}^{(4)}&\mathbf{a}_{222}^{(4)}
\end{pmatrix}.
\end{align*}
More examples can be found in~\cite{kolda2009tensor}.
\begin{definition}
\label{def:mode-nproduck}
Let $\mathcal{A} \in \mathbb{R}^{m_1\times m_2\times....\times m_n}$ an order-\textit{n} tensor and $\mathbf{B} \in \mathbb{R}^{d\times m_k}$ a matrix Then the mode-$k$ product $\mathcal{A} \times_k \mathbf{B}$  is defined as:
\[
\mathcal{A} \times_k \mathbf{B} = \texttt{fold}^{(\texttt{k})}\big(\mathbf{B}\cdot   \texttt{unfold}^{(\texttt{k})}(\mathcal{A})\big).
\]
\end{definition}

\begin{definition}
\label{def:facewise}
The facewise product multiplies each of the frontal slices of two tensors. Let $\mathcal{A} \in \mathbb{R}^{m_1\times \ell \times....\times m_n}$ and $\mathcal{B} \in \mathbb{R}^{\ell \times m_2\times....\times m_n}$ be order-\textit{n} tensors. Then the facewise product $\mathcal{C} =\mathcal{A} \Delta \mathcal{B} \in \mathbb{R}^{m_1 \times m_2\times....\times m_n}$ is defined as:
\begin{align*}
\mathcal{C}_{(i_3i_4...i_n)} &= \mathcal{A}_{(i_3i_4...i_n)} \cdot \mathcal{B}_{(i_3i_4...i_n)},  
\end{align*}
for $i_k=1,...,m_k$, where $k = 3,...,n$.

\end{definition}

\begin{definition}
\label{def:fro}
The tensor norm used through this paper is the Frobenious norm which for the tensor $\mathcal{A} \in \mathbb{R}^{m_1\times m_2\times....\times m_n}$ is given by:
\begin{equation*}
\label{eq:fro}
    ||\mathcal{A}||_F = \sqrt{\sum_{i=1}^{m_1}\sum_{j=1}^{m_2}...\sum_{k=1}^{m_n}\big(\mathcal{A}(i,j,...,k)}\big)^2.
\end{equation*}
\end{definition}
\subsection{Tensor operators}
Recently, a family of tensor-tensor products for third-order tensors has been formulated in a so-called transform domain for any invertible linear transform~\cite{kernfeld2015tensor}. In this subsection, we extend this development so that the computation of order-\textit{n} tensor operators can be more easily defined (and computed) in the transform domain rather than the spatial domain.  In order to generalize the order-n tensor operators, we use “$L$” subscript
which refers to any invertible linear transformation.
\label{subsec:tensoroperators}
\begin{definition}
\label{def:tensorproduct}
Let $\mathcal{A} \in \mathbb{R}^{m_1\times \ell \times....\times m_n}$ and $\mathcal{B} \in \mathbb{R}^{\ell \times m_2\times....\times m_n}$ be order-\textit{n} tensors. The tensor-tensor product based on $L$ transform $\mathcal{A}*_L\mathcal{B} \in \mathbb{R}^{m_1\times m_2 \times....\times m_n}$ is defined as:
\begin{align*}
    &\Tilde{\mathcal{A}} = \mathcal{A} \times_3 \mathbf{L}_3 \times_4 \mathbf{L}_4 \times...\times_n \mathbf{L}_n,\\
    &\Tilde{\mathcal{B}} = \mathcal{B} \times_3 \mathbf{L}_3 \times_4 \mathbf{L}_4 \times...\times_n \mathbf{L}_n,\\
    &\mathcal{A}*_L\mathcal{B} = (\Tilde{\mathcal{A}} \Delta \Tilde{\mathcal{B}}) \times_3 \mathbf{L}_3^{-1} \times...\times_n \mathbf{L}_n^{-1},\\
\end{align*}
\end{definition}
\noindent where $\mathbf{L}_i$ is an $m_i \times m_i$ invertible transformation matrix where $i=1,...,n$.

\begin{algorithm}[H]
\caption{tensor-tensor product induced by $*_L$}\label{alg:tprod}
\begin{algorithmic}
\STATE 
\STATE {\textbf{Input:}} Input tensors $\mathcal{A} \in \mathbb{R}^{m_1 \times \ell \times...\times m_n}$ and $\mathcal{B} \in \mathbb{R}^{\ell \times m_2 \times...\times m_n}$
\STATE {\textbf{Output:}} $\mathcal{C} \in \mathbb{R}^{m_1 \times m_2 \times...\times m_n}$
\FOR{$i=3$ to $n$} 
\STATE $\tilde{\mathcal{A}} \leftarrow L\big(\mathcal{A},[\;],i\big)$
\STATE $\tilde{\mathcal{B}} \leftarrow L\big(\mathcal{B},[\;],i\big)$
\ENDFOR
\FOR{$i=1$ to $m_3$}
\STATE $\vdots$
\FOR{$k=1$ to $m_n$ }
\STATE $\tilde{\mathcal{C}}(:,:,i,..,k) =\tilde{\mathcal{A}}(:,:,i,...,k)\times \tilde{\mathcal{B}}(:,:,i,...,k)$
\ENDFOR
\STATE $\vdots$
\ENDFOR
\FOR{$i=n$ to $3$}
\STATE $\mathcal{C} \leftarrow L^{-1}\big(\tilde{\mathcal{C}},[\;],i\big)$
\ENDFOR
\end{algorithmic}
\end{algorithm}

\begin{definition}
\label{def:identity}
The identity tensor
$\mathcal{I} \in \mathbb{R}^{m \times m \times m_3 \times...\times m_n}$ is the tensor whose frontal slice is the $m \times m$ identity
matrix in the transform domain.
\[
\mathcal{I} = \Tilde{\mathcal{I}} \times_3 \mathbf{L}_3^{-1} \times_4 \mathbf{L}_4^{-1} \times...\times_n \mathbf{L}_n^{-1},
\]
where $\Tilde{\mathcal{I}}(:,:,i_3,i_4,...,i_n) = \mathbf{I}$ for $i_k=1,...,m_k$ and $k = 3,...,n$ and $\mathbf{I}$ is the $m \times m$ identity matrix.
\end{definition}

\begin{definition}
\label{def:inv}
A tensor $\mathcal{A} \in \mathbb{R}^{m \times m \times m_3 \times...\times m_4}$ has an tensor inverse $\mathcal{B} \in \mathbb{R}^{m \times m \times m_3 \times...\times m_n}$ provided:
$$
\mathcal{A}*_L\mathcal{B} = \mathcal{I}\; \text{and}\; \mathcal{B}*_L\mathcal{A} = \mathcal{I},
$$
where $\mathcal{I} \in \mathbb{R}^{m \times m \times m_3 \times...\times m_n}$.  The tensor inverse is computed as:  
\begin{align*}
&\mathcal{B} = \texttt{inv}_L(\mathcal{A}),\\
&\Tilde{\mathcal{A}} = \mathcal{A} \times_3 \mathbf{L}_3 \times_4 \mathbf{L}_4 \times...\times_n \mathbf{L}_n,\\
&\mathcal{B}_{(i_3i_4...i_n)} = (\Tilde{\mathcal{A}}_{(i_3i_4...i_n)})^{-1},
\end{align*}
for $i_k=1,...,m_k$, where $k = 3,...,n$.
\end{definition}

\begin{algorithm}[H]
\caption{{tensor inverse (\texttt{inv$_L$}})}\label{alg:inv}
\begin{algorithmic}
\STATE 
\STATE {\textbf{Input:}} Input tensors $\mathcal{A} \in \mathbb{R}^{m \times m \times m_3 \times...\times m_n}$
\STATE {\textbf{Output:}} $\mathcal{B} \in \mathbb{R}^{m \times m \times m_3 \times...\times m_n}$
\FOR{$i=3$ to $n$} 
\STATE $\tilde{\mathcal{A}} \leftarrow L\big(\mathcal{A},[\;],i\big)$
\ENDFOR
\FOR{$i=1$ to $m_3$}
\STATE $\vdots$
\FOR{$k=1$ to $m_n$ }
\STATE $\tilde{\mathcal{B}}(:,:,i,..,k) =\tilde{\mathcal{A}}(:,:,i,...,k)^{-1}$
\ENDFOR
\STATE $\vdots$
\ENDFOR
\FOR{$i=n$ to $3$}
\STATE $\mathcal{B} \leftarrow L^{-1}\big(\tilde{\mathcal{B}},[\;],i\big)$
\ENDFOR
\end{algorithmic}
\end{algorithm}

\begin{definition}
\label{def:trans}
If $\mathcal{A} \in \mathbb{R}^{m_1\times m_2\times....\times m_n}$, then the tensor transpose $\texttt{trans$_L$}(\mathcal{A}) \in \mathbb{R}^{m_2\times m_1\times....\times m_n}$ is defined as:
\begin{align*}
    &\mathcal{B} = \texttt{trans}_L(\mathcal{A}),\\
    &\Tilde{\mathcal{A}} = \mathcal{A} \times_3 \mathbf{L}_3 \times_4 \mathbf{L}_4 \times...\times_n \mathbf{L}_n,\\
    &\Tilde{\mathcal{B}}_{(i_3i_4...i_n)} = (\Tilde{\mathcal{A}}_{(i_3i_4...i_n)})^{T},\\
    &\mathcal{B} = \Tilde{\mathcal{B}} \times_3 \mathbf{L}_3^{-1} \times...\times_n \mathbf{L}_n^{-1},
\end{align*}
for $i_k=1,...,m_k$, where $k = 3,...,n$.
\end{definition}

\begin{algorithm}[H]
\caption{tensor transpose (\texttt{trans$_L$})}\label{alg:trans}
\begin{algorithmic}
\STATE 
\STATE {\textbf{Input:}} Input tensors $\mathcal{A} \in \mathbb{R}^{m_1 \times m_2 \times...\times m_n}$
\STATE {\textbf{Output:}} $\mathcal{B} \in \mathbb{R}^{m_2 \times m_1 \times...\times m_n}$
\FOR{$i=3$ to $n$} 
\STATE $\tilde{\mathcal{A}} \leftarrow L\big(\mathcal{A},[\;],i\big)$
\ENDFOR
\FOR{$i=1$ to $m_3$}
\STATE $\vdots$
\FOR{$k=1$ to $m_n$ }
\STATE $\tilde{\mathcal{B}}(:,:,i,..,k) =\tilde{\mathcal{A}}(:,:,i,...,k)^T$
\ENDFOR
\STATE $\vdots$
\ENDFOR
\FOR{$i=n$ to $3$}
\STATE $\mathcal{B} \leftarrow L^{-1}\big(\tilde{\mathcal{B}},[\;],i\big)$
\ENDFOR
\end{algorithmic}
\end{algorithm}

\subsection{Order-\textit{n} Tensor Eigendecomposition}
\label{subsec:t-eig}
The final tool necessary for a multilinear LDA is to define a tensor-tensor eigenvalue decomposition. In our previous work, tensor-tensor eigendecomposition has been introduced for third order tensor based upon on the discrete Fourier transform (DFT)~\cite{hoover2018multilinear}. In this subsection, we extend this work to order-\textit{n} tensor by utilization of the new tensor operators defined in \textbf{Section}~\ref{subsec:tensoroperators}. 
\begin{definition}
\label{def:matrixeig}
Let $\mathbf{A} \in \mathbb{R}^{m \times m}$ be a matrix. Then the eigenvalue decomposition is given as:
\[
\mathbf{A} = \mathbf{Q}\mathbf{\Lambda}\mathbf{Q}^{-1}
\]
\end{definition}
\noindent where $\mathbf{Q} \in \mathbb{R}^{m \times m}$ is an orthogonal matrix and $\mathbf{\Lambda} \in \mathbb{R}^{m \times m}$ is a diagonal matrix.

\begin{figure*}
    \centering
    \includegraphics[width=0.8\textwidth]{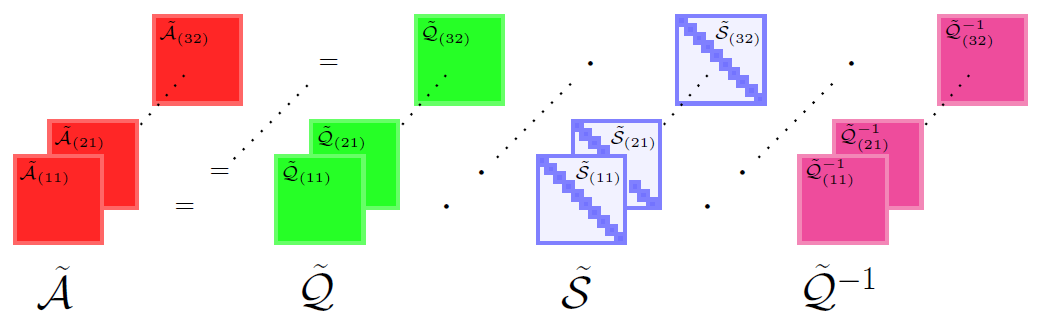}
    \caption{Illustration of the computation of the tensor eigendecomposition in the transform domain for $\mathcal{A} \in \mathbb{R}^{m \times m \times 3 \times 2}$.}
    \label{fig:teig}
\end{figure*}

\begin{definition}
\label{def:t-eig}
If $\mathcal{A} \in \mathbb{R}^{m\times m\times m_3 \times....\times m_n}$, then the tensor-tensor eigendecomposition \texttt{t-eig$_{L}$} is formed by computing the matrix eigenvalue decomposition for the frontal slices of $\mathcal{A}$ in the transform domain. 
\begin{align*}
     &\Tilde{\mathcal{A}} = \mathcal{A} \times_3 \mathbf{L}_3 \times_4 \mathbf{L}_4 \times...\times_n \mathbf{L}_n,\\
     &\Tilde{\mathcal{A}}_{(i_1i_2...i_n)} = \Tilde{\mathcal{Q}}_{(i_1i_2...i_n)} \cdot \Tilde{\mathcal{S}}_{(i_1i_2...i_n)} \cdot \Tilde{\mathcal{Q}}^{-1}_{(i_1i_2...i_n)},\\
&\text{for $i_k = 1,...,m_k$, where $k = 1,...,n$ },\\
     & {\mathcal{Q}} = \tilde{\mathcal{Q}} \times_3 \mathbf{L}^{-1}_3 \times_4 \mathbf{L}^{-1}_4 \times...\times_n \mathbf{L}^{-1}_n,\\
     & {\mathcal{S}} = \tilde{\mathcal{S}} \times_3 \mathbf{L}^{-1}_3 \times_4 \mathbf{L}^{-1}_4 \times...\times_n \mathbf{L}^{-1}_n.
\end{align*}
Therefore;
\begin{align*}
	&\mathcal{Q},\mathcal{S} = \texttt{t-eig}_{L}(\mathcal{A}),\\
    &\mathcal{A} = \mathcal{Q}*_L\mathcal{S}*_L\texttt{inv$_L$}(\mathcal{Q}),
\end{align*}
where $\mathcal{Q} \in \mathbb{R}^{m \times m \times m3 \times...\times m_n}$ is an orthogonal tensor such that $\mathcal{Q}*_L\texttt{trans$_L$}(\mathcal{Q})=\mathcal{I}$ and $\texttt{trans$_L$}(\mathcal{Q})*_L\mathcal{Q}=\mathcal{I}$ and $\mathcal{S} \in \mathbb{R}^{m \times m \times m3 \times...\times m_n}$ is a diagonal tensor such that the frontal slices are diagonal. A graphical illustration of the computation of the tensor eigendecomposition in the transform domain is shown in Fig.~\ref{fig:teig}.

\end{definition}

Similar \textbf{Algorithms~\ref{alg:tprod}-\ref{alg:trans}} above, we provide the algorithmic pseudo-code for computing the {\texttt{t-eig$_{L}$}},  below.

\begin{algorithm}[H]
\caption{tensor eigendecomposition ({\texttt{t-eig$_{L}$}})}\label{alg:t-eig}
\begin{algorithmic}
\STATE 
\STATE {\textbf{Input:}} Input tensors $\mathcal{A} \in \mathbb{R}^{n \times n\times m_3 \times...\times m_n}$ 
\STATE {\textbf{Output:}} $\mathcal{Q} \in \mathbb{R}^{n \times n \times m_3 \times...\times m_n}$, $\mathcal{S} \in \mathbb{R}^{n \times n \times m_3 \times...\times m_n}$
\FOR{$i=3$ to $n$} 
\STATE $\tilde{\mathcal{A}} \leftarrow L\big(\mathcal{A},[\;],i\big)$
\ENDFOR
\FOR{$i=1$ to $m_3$}
\STATE $\vdots$
\FOR{$k=1$ to $m_n$ }
\STATE $Q,S=$ \textit{eig}\big($\tilde{\mathcal{A}}(:,:,i,...,k)$\big)
\STATE $\tilde{\mathcal{Q}}(:,:,i,...,k)=Q$
\STATE $\tilde{\mathcal{S}}(:,:,i,...,k)=S$
\ENDFOR
\STATE $\vdots$
\ENDFOR
\FOR{$i=n$ to $3$}
\STATE $\mathcal{Q} \leftarrow L^{-1}\big(\tilde{\mathcal{Q}},[\;],i\big)$
\STATE $\mathcal{S} \leftarrow L^{-1}\big(\tilde{\mathcal{S}},[\;],i\big)$
\ENDFOR
\end{algorithmic}
\end{algorithm}

\section{High-Order Multilinear Discriminant Analysis}
\label{sec:MLDA}
\subsection{ Linear Discriminant Analysis}
  The general idea behind LDA is to compute a projection matrix $U$ that maximizes the between class means while simultaneously minimizing the within class means.  Such a projection is computed by deriving two scatter matrices to account for variation both within and between different classes.  We consider a set of data samples that contain $c$ classes (with associated class labels) and each class contains $n_i$ data samples, i.e., $i = 1, 2, \dots, c$.  Denoting each class as $c_i$ we construct the within-class scatter matrix as:
\begin{equation}
	\mathbf{S}_W = \sum_{i = 1}^c \mathbf{S}_i,
\end{equation}
where 
\begin{equation}
	\mathbf{S}_i = \sum_{\textbf{x} \in c_i} (\textbf{x} - \textbf{m}_i)(\textbf{x} - \textbf{m}_i)^T,
\end{equation}
$\textbf{x} \in \mathbb{R}^m$ is an $m$-dimensional data sample, and $\textbf{m}_i$ is the mean of class $i$, i.e.,
\begin{equation}
	\textbf{m}_i = \frac{1}{n_i} \sum_{\textbf{x} \in c_i} \textbf{x},
    \label{eq:class_mean}
\end{equation}
where $n_i$ the total number of samples in the class $i$.

The between-class scatter matrix is constructed to account for the class means around the total mean of the data as:
\begin{equation}
	\mathbf{S}_B = \sum_{i = 1}^c n_i (\textbf{m}_i - \textbf{m})(\textbf{m}_i - \textbf{m})^T,
\end{equation}
where $\textbf{m}$ is the mean of all data samples, and $\textbf{m}_i$ is defined in~(\ref{eq:class_mean}).  The projection matrix $U$ can then be computed by maximizing the ratio of determinants between $S_W$ and $S_B$ in the projection space as:
\begin{equation}
	\argmax_\mathbf{U} \frac{|\mathbf{U}^T \mathbf{S}_B \mathbf{U}|}{|\mathbf{U}^T \mathbf{S}_W \mathbf{U}|}.
    \label{eq:LDA}
\end{equation}
Re-casting~(\ref{eq:LDA}) as a constrained optimization problem, it can be shown that the solution is computed by solving the generalized eigenvalue problem.
\begin{equation}
	\mathbf{S}_B \textbf{u}_p = \lambda_p \mathbf{S}_W \textbf{u}_p,
    \label{eq:gen_eig}
\end{equation}
where $\mathbf{U} = [\textbf{u}_1,\textbf{u}_2, \dots, \textbf{u}_p]$ corresponds to the $p$ largest eigenvalues $\lambda_p$~\cite{hart2000pattern,belhumeur1997eigenfaces,fisher1936use}.  Note that there are at most $c-1$ nonzero eigenvalues of~(\ref{eq:gen_eig}) therefore the projection space has at most dimension $c-1$.

\subsection{High-Order Multilinear Discriminant Analysis}
When LDA is used for higher-order databases, vectorizing all data samples destroys the spatial correlation within each sample and may cause small sample size problem~\cite{fukunaga2013introduction}. Even though the MDA methods  address these issues, all the MDA methods are constructed using the \textit{n}-mode product and Tucker decomposition structure~\cite{yan2005discriminant,tao2007general,li2014multilinear,de2000multilinear}. Alternative to the MDA methods, we propose a new approach referred to as high-order multilinear discriminant analysis (HOMLDA) built upon the order-\textit{n} tensor operators and the order-\textit{n} tensor eigendecomposition defined in Section~\ref{subsec:tensoroperators} and Section~\ref{subsec:t-eig} respectively . We can represent a higher-order data as an order-\textit{n} tensor by stacking all the tensor samples as lateral slices into a tensor structure.

Suppose we have $\ell$ number of input samples where each sample is a $m_1 \times m_2 \times...\times m_n$ tensor. Then the data tensor $\mathcal{A} \in \mathbb{R}^{m_1 \times \ell \times m_2 \times...\times m_n}$ is constructed by stacking all the input samples as lateral slices into the tensor $\mathcal{A}$.
The within-class tensor can be computed as:
\begin{equation}
    \mathcal{W} = \sum_{i=1}^{c} \sum_{\Vec{\mathcal{A}}_{(j)}\in \text{c}_i} (\Vec{\mathcal{A}}_{(j)} - \Vec{\mathcal{M}}_{i}) *_L \texttt{trans$_L$}(\Vec{\mathcal{A}}_{(j)}- \Vec{\mathcal{M}}_{i}),
    \label{eq:W}
\end{equation}
where the tensor operators ``$*_L$" and ``$\texttt{trans$_L$}$" are given in \textbf{Definition~\ref{def:tensorproduct}} and \textbf{Definition~\ref{def:trans}} respectively. $c$ is the total number of classes, $\Vec{\mathcal{M}}_{i}$ is the mean tensor corresponding to the class $i$.
\begin{equation}
    \Vec{\mathcal{M}}_{i} = \frac{1}{n_i}\sum_{\Vec{\mathcal{A}}_{(j)}\in \text{c}_i}{\Vec{\mathcal{A}}_{(j)}},
\end{equation}
where $n_i$ the total number of samples in the class $i$. We remind the reader that $\Vec{\mathcal{A}}_{(j)}$ is the $j^{\text{th}}$ lateral slice of the tensor $\mathcal{A}$.

We define the between-class scatter tensor as:
\begin{equation}
\label{eq:B}
    \mathcal{B} = \sum_{i=1}^{c}n_i(\Vec{\mathcal{M}}_{i}-\Vec{\mathcal{M}})*_L\texttt{trans$_L$}(\Vec{\mathcal{M}}_{i}-\Vec{\mathcal{M}}),
\end{equation}
where $\Vec{\mathcal{M}}$ is the mean of all data samples.

The projection tensor $\mathcal{U}$ can then be computed by solving the generalized tensor eigenvalue problem as:
\begin{equation}
\label{eq:t-eig}
	\big( \texttt{inv$_L$}(\mathcal{W}) *_L \mathcal{B} \big) *_L \mathcal{U} = \mathcal{U} *_L \mathcal{S},
\end{equation}
The projection tensor $\mathcal{U}_p = [\Vec{\mathcal{U}}_1,\Vec{\mathcal{U}}_2, \dots, \Vec{\mathcal{U}}_p] \in \mathbb{R}^{m_1 \times p \times m_2 \times...\times m_n}$ consists of the eigentensors (the lateral slices of the tensor $\mathcal{U}$)  corresponding to the $p$ largest eigentuples of the diagonal tensor $\mathcal{S} \in \mathbb{R}^{p \times p \times m_2 \times...\times m_n}$. Note that similar to its matrix counterpart, there are at most $c-1$ nonzero eigentuples of~(\ref{eq:t-eig}).  In addition, because the within- and between-class scatter tensors are of size $m_1 \times m_1 \times m_2 \times...\times m_n$, the small sample size issue is non-existent and the computation of $\mathcal{S}$ and $\mathcal{U}$ can be performed via utilization of the $\texttt{t-eig}_{L}$ operator defined in \textbf{Definition~\ref{def:t-eig}} as:
\begin{equation}
	\mathcal{S},\mathcal{U} = \texttt{t-eig}_{L}\big( \texttt{inv$_L$}(\mathcal{W}) *_L \mathcal{B} \big)
\end{equation}

We can project the data tensor $\mathcal{A}$ onto the low-dimensional subspace $\mathcal{U}_p$ via:
\begin{equation}
   \mathcal{T} =\texttt{trans}_L(\mathcal{U}_p) *_L \mathcal{A}, 
\end{equation}
where $\mathcal{T} \in \mathbb{R}^{p \times \ell \times m_3 \times...\times m_n}$.

\subsection{Robust High-Order Multilinear Discriminant Analysis}
In our proposed method HOMLDA, the within-class and between-class scatter tensors are computed as given in~(\ref{eq:W})
and~(\ref{eq:B}) respectively. The projection tensor is also computed by solving the generalized tensor eigenvalue problem as given~(\ref{eq:t-eig}). It can be seen that the tensor inverse (defined in \textbf{Definition~\ref{def:inv}} and \textbf{Algorithm~\ref{alg:inv}}) of the within-class scatter tensor  needs to be calculated. However, the within-class scatter tensor might be close to singularity. This might cause the calculation of the tensor inverse of the within-class scatter tensor is not accurate and lead to distort HOMLDA. Recently, a Robust Linear Discriminant Analysis (RLDA) has been introduced to address the singularity problem of the within-class scatter matrix of the classic LDA~\cite{guo2015feature}. In this subsection, we provide a multilinear extension of RLDA and propose a novel Robust High-Dimensional Multilinear Discriminant
Analysis (RHOMLDA) that improves the generalization capability of HOMLDA by the robust estimate of the inverse of the within-class scatter tensor.

Suppose we have a within-class scatter tensor $\mathcal{W} \in \mathbb{R}^{m_1 \times m_2 \times m_3...\times m_n}$. We note that $m_1$ must be equal to $m_2$, as the frontal slices of a within-class scatter tensor are always square matrices.
Since the tensor inverse operator computes the matrix inverse of all the frontal slices of $\mathcal{W}$ in the transform domain, we apply a reconstruction for the frontal slices which are ill-conditioned; whereas we keep the frontal slices which are well-conditioned. In order to determine which frontal slices of the tensor $\mathcal{W}$ are ill-conditioned, the condition numbers needs to be calculated for each frontal slice in the transform domain.
\begin{align}
    &\Tilde{\mathcal{W}} = \mathcal{W} \times_3 \mathbf{L}_3 \times_4 \mathbf{L}_4 \times...\times_n \mathbf{L}_n, \nonumber\\
    &\kappa(\Tilde{\mathcal{W}}_{(i_3i_4...i_n)}) = ||\Tilde{\mathcal{W}}_{(i_3i_4...i_n)}||_F ||\Tilde{\mathcal{W}}_{(i_3i_4...i_n)}^{-1}||_F,\nonumber\\
     &\text{for $i_k = 1,...,m_k$, where $k = 1,...,n$ },
\end{align}
where $\kappa(\Tilde{\mathcal{W}}_{(i_3i_4...i_n)})$ denotes the condition number of the corresponding frontal slice and $||.||$ is the tensor norm defined in \textbf{Definition~\ref{def:fro}}. In this paper, we use $10^5$ as the threshold condition number. Therefore, the ill-conditioned frontal slices whose condition numbers are greater than $10^5$ need to be reconstructed.

First, we factorize the within-class scatter tensor $\mathcal{W}$ defined in \textbf{Definition~\ref{def:t-eig}} and \textbf{Algorithm~\ref{alg:t-eig}} as:
\begin{align}
  	&\mathcal{Q},\mathcal{S} = \texttt{t-eig}_{L}(\mathcal{W}),\nonumber\\
    &\mathcal{W} = \mathcal{Q}*_L\mathcal{S}*_L\texttt{inv$_L$}(\mathcal{Q}).  
\end{align}
Therefore, an ill-condition frontal slice can be written as:
\begin{align}
        &\Tilde{\mathcal{W}}_{(i_1i_2...i_n)} = \Tilde{\mathcal{Q}}_{(i_1i_2...i_n)} \cdot \Tilde{\mathcal{S}}_{(i_1i_2...i_n)} \cdot \Tilde{\mathcal{Q}}^{-1}_{(i_1i_2...i_n)}, 
\end{align}
where $\Tilde{\mathcal{Q}}_{(i_1i_2...i_n)}$ is an orthogonal matrix, and $\Tilde{\mathcal{S}}_{(i_1i_2...i_n)}$ is a diagonal matrix with the eigenvalues in decreasing order ($\lambda_1 \geq \lambda_2 \geq... \lambda_{p}$). We assume that the first $k$ eigenvalues are reliable, and the rest need to be re-estimated. We use the weighted average method~\cite{moghaddam1997probabilistic,deng2007robust} to re-estimate the remaining $p-k$ eigenvalues. The estimation for the  the remaining $p-k$ eigenvalues can be written as:
\begin{equation}
    \lambda^*=\frac{1}{p-k}\sum_{i=k+1}^{p}\lambda_i.
\end{equation}
Thus, we re-estimate an ill-condition frontal slice of the tensor $\mathcal{W}$ as:
\begin{align}
    &\Tilde{\Hat{\mathcal{W}}}_{(i_1i_2...i_n)} = \Tilde{\mathcal{Q}}_{(i_1i_2...i_n)} \cdot \Tilde{\Hat{\mathcal{S}}}_{(i_1i_2...i_n)} \cdot \Tilde{\mathcal{Q}}^{-1}_{(i_1i_2...i_n)},\nonumber\\
    &\Tilde{\Hat{\mathcal{S}}}_{(i_1i_2...i_n)} = diag(\lambda_1, \lambda_2,..., \lambda_{k},\lambda^*,...,\lambda^*).
\end{align}
To determine $k$, the number of reliable eigenvalues, we can use the following formula:
\begin{equation}
    \arg \min_{k}\frac{\sum_{i=1}^{k}\lambda_i}{\sum_{i=1}^{i=p}\lambda_i} \geq E(m),
\end{equation}
which represents the proportion of reliable energy encoded in the first $k$ eigenvalues. We set $E(m) = 0.98$ for our experiments in this paper. Finally, the new within-class scatter tensor $\Hat{\mathcal{W}}$ can be computed as:
\begin{equation}
    \Hat{\mathcal{W}} = \mathcal{Q}*_L\Hat{\mathcal{S}}*_L\texttt{inv$_L$}(\mathcal{Q}).  
\end{equation}
We remind the readers that we only re-estimate the ill-conditioned frontal slices of $\mathcal{W}$, whereas, we keep the well-conditioned frontal slices of $\mathcal{W}$.

\textbf{Algorithm~\ref{alg:update_W}} provides the pseudo-code for the new within-class scatter tensor.

\begin{algorithm}[H]
\caption{update the within-class scatter tensor $\mathcal{W}$}
\label{alg:update_W}
\begin{algorithmic}
\STATE 
\STATE {\textbf{Input:}} Input tensors $\mathcal{W} \in \mathbb{R}^{m_1 \times m_2\times...\times m_n}$ 
\vspace{1pt}
\STATE {\textbf{Output:}}$\Hat{\mathcal{W}} \in \mathbb{R}^{m_1 \times m_2\times...\times m_n}$ 
\FOR{$i=3$ to $n$} 
\STATE $\tilde{\mathcal{W}} \leftarrow L\big(\mathcal{W},[\;],i\big)$
\ENDFOR
\FOR{$i=1$ to $m_3$}
\STATE $\vdots$
\FOR{$k=1$ to $m_n$ }
\STATE $\kappa =||\Tilde{\mathcal{W}}(:,:,i,...,k)||_F ||\Tilde{\mathcal{W}}^{-1}(:,:,i,...,k)||_F$
\IF{$\kappa \geq 10^5$}
\STATE $\Tilde{\mathcal{W}}(:,:,i,...,k) = $
\STATE $\Tilde{\mathcal{Q}}(:,:,i,...,k) \cdot \Tilde{\mathcal{S}}(:,:,i,...,k) \cdot \Tilde{\mathcal{Q}}^{-1}(:,:,i,...,k)$
\STATE $\Tilde{\mathcal{S}}(:,:,i,...,k) = diag(\lambda_1,\lambda_2,...,\lambda_p)$
\STATE $\argminl_{k} \frac{\sum_{i=1}^{k}\lambda_i}{\sum_{i=1}^{i=p}\lambda_i} \geq 0.98$
\STATE $\lambda^*=\frac{1}{p-k}\sum_{i=k+1}^{p}\lambda_i$
\vspace{1pt}
\STATE $\Tilde{\Hat{\mathcal{S}}}(:,:,i,...,k) = diag(\lambda_1, \lambda_2,..., \lambda_{k},\lambda^*,...,\lambda^*)$
\vspace{1pt}
\STATE $\Tilde{\Hat{\mathcal{W}}}(:,:,i,...,k) = $
\STATE $\Tilde{\mathcal{Q}}(:,:,i,...,k) \cdot \Tilde{\Hat{\mathcal{S}}}(:,:,i,...,k) \cdot \Tilde{\mathcal{Q}}^{-1}(:,:,i,...,k)$
\ELSE
\vspace{1.5pt}
\STATE $\Tilde{\Hat{\mathcal{W}}}(:,:,i,...,k) = \Tilde{\mathcal{W}}(:,:,i,...,k)$
\ENDIF
\ENDFOR
\STATE $\vdots$
\ENDFOR
\FOR{$i=n$ to $3$}
\vspace{1.5pt}
\STATE ${\Hat{\mathcal{W}}} \leftarrow L^{-1}\big(\Tilde{\Hat{\mathcal{W}}},[\;],i\big)$
\ENDFOR
\end{algorithmic}
\end{algorithm}
\label{subsec:RHOMLDA}

\section{experimental results}
\label{experimental}
In this section we compare our proposed method with Tucker structure based multilinear discriminant methods, CMDA and DGTDA~\cite{li2014multilinear}. We use three different invertible linear transformations for our proposed HOMLDA and RHOMLDA approaches, namely the discrete Fourier transform (DFT), the discrete cosine transform (DCT), and the discrete wavelet transform (DWT).
Unlike the DFT or DCT, there are different types of wavelets functions that can be used to define the wavelet basis~\cite{strang1996wavelets}. Most notably are the Haar wavelet~\cite{haar1910theorie,porwik2004haar} and Daubechies wavelet~\cite{daubechies1993orthonormal}. In our work, we use the level-1 Haar discrete wavelet transformation matrix due to their low computational cost and simplicity to apply as compared to other wavelets.
Although any invertible linear transformation can be used for both HOMLDA and RHOMLDA, the recognition performance of one could be better than others for a particular data set.  

The experiments were conducted on four different data sets: the FEI face dataset~\cite{thomaz2010new}, the Multimedia University (MMU) iris database~\cite{web:MMU}, the Pattern Recognition and Digital Image Processing Group of the University of Pernambuco (RPPDI) dynamic gestures database~\cite{barros2017dynamic}, and the University of Texas at Dallas Multimodal Human Action Dataset (UTD-MHAD)~\cite{chen2015utd}.  Classification accuracy is evaluated using a nearest neighbor search.
\subsection{FEI Face Database}
\begin{figure}[!h]
    \centering
    \includegraphics[width=\columnwidth]{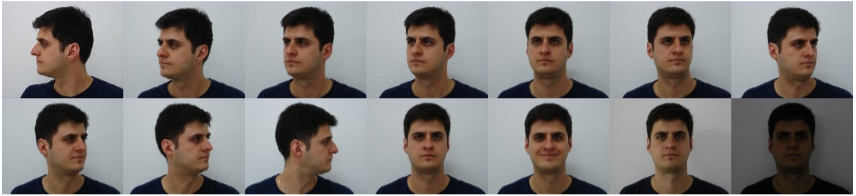}
    \caption{All $14$ images of an individual in the FEI face database.}
    \label{fig:fei}
\end{figure}

The FEI face database contains a set of RGB face images taken at the Artificial Intelligence Laboratory of FEI~\cite{thomaz2010new}. There are $14$ images for each of $200$ individuals, a total of 2800 images. All images are taken against a white homogeneous background in an upright frontal position with profile rotation of up to about 180 degrees. All $14$ images of an individual is illustrated in Fig.~\ref{fig:fei}. The original size of each image is $480\times640$. For computational efficiency,  we downsampled each image to $48\times64$. The data set\footnote{As the length of a signal must be 2m for the level-1 Haar wavelet transform, we implement zero padding on the mode-4 to turn the dimension into a length of 2m. Therefore, we add one more dimension with all zeros. The data set we used for the DWT is a tensor $\mathcal{Y} \in \mathbb{R}^{ 48\times2800\times64\times4}$.} is represented as a tensor  $\mathcal{Y} \in \mathbb{R}^{ 48\times2800\times64\times3}$. 

\subsection{MMU Irish Database}
\begin{figure}[!h]
    \centering
    \includegraphics[width=\columnwidth]{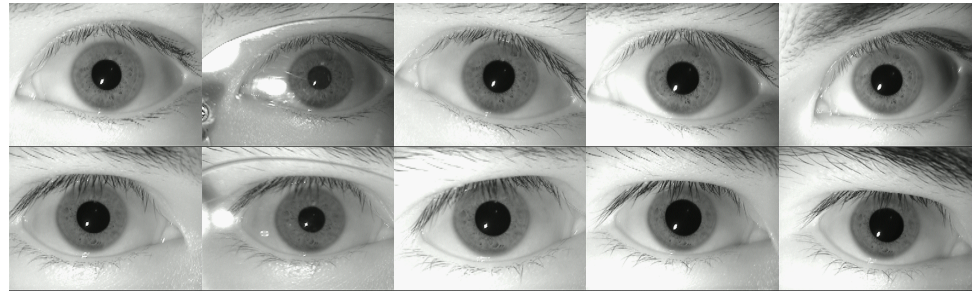}
    \caption{Left and right iris images of an individual}
    \label{fig:coil}
\end{figure}

The MMU data set includes eye images of $45$ different objects.
There are both $5$ images of left and right iris of $45$ people, totaling $450$ irish images. Each image is a 24-bit image consists of three channels of 8-bit images. Fig.~\ref{fig:coil} shows left and right iris images of an indiviual on the first-row and the second-row respectively. All images of size $240 \times 320$ are  downsampled to $24 \times 32$. Each person is repented as a 
 tensor $\mathcal{X} \in \mathbb{R}^{ 24\times1\times32\times6}$. The first and the third-mode represent row and column pixels, respectively, whereas the fourth-mode represents left and right iris images of a person. Hence, the data set is represented as a tensor  $\mathcal{Y} \in \mathbb{R}^{ 24\times225\times32\times6}$.

\subsection{RPPDI Dynamic Gestures Database}
\begin{figure}[h!]
    \centering
    \includegraphics[width=\columnwidth]{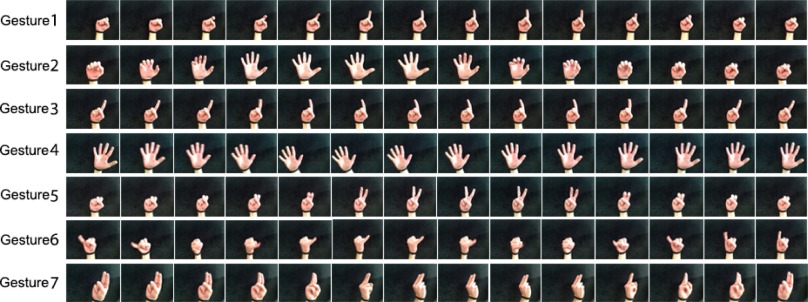}
    \caption{A subset of the samples of the RPPDI hand gesture database from $7$ different gesture classes.}
    \label{fig:chg}
    \vspace{-3ex}
\end{figure}

The RPPDI dynamic gestures database consists of $188$ video sequences of $7$ gesture classes. For each video sample, there are $14$ image sequences consisting of RGB images of size $480\times640$. Fig.~\ref{fig:chg} illustrates a subset of the samples for each class.  In our experiments, all video sequences are grayscaled and resized to $48\times64$. Thus, the data can be represented as a tensor  $\mathcal{Y} \in \mathbb{R}^{ 48\times188\times64\times 14}$.

\subsection{UTD-MHAD Database}
\begin{figure}[!h]
    \centering
    \includegraphics[width=\columnwidth]{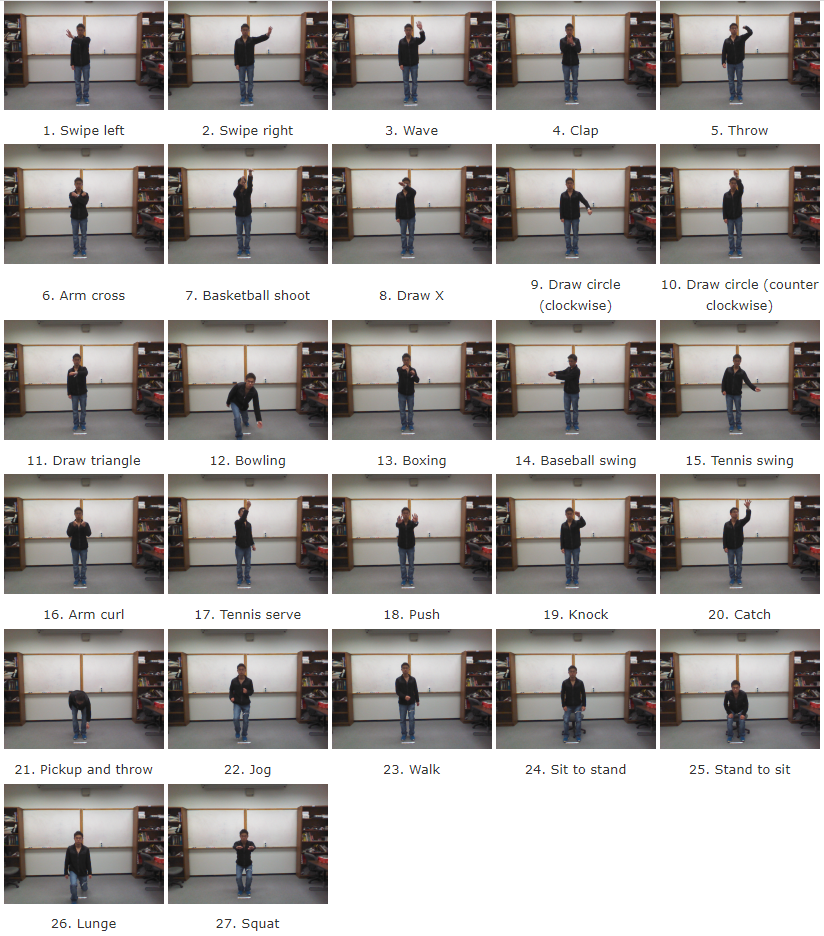}
    \caption{Example frames of the UTD-MHAD database from $27$ different actions. }
    \label{fig:ha}
\end{figure}

The UTD-MHAD database contains $27$ actions performed by $8$ subjects. Fig.~\ref{fig:ha} shows a frame from each action. Each subject repeated each action $4$ times. Thus, there are $32$ videos for each class and sequence length varies.  For our experiment, each frame are grayscaled and resized from $480 \times 640$ to $24 \times 32$ and the middle $32$ frames are obtained to balance the class size. Thus, we construct a tensor $\mathcal{Y} \in \mathbb{R}^{ 24\times864\times32\times 32}$.

\subsection{Classification accuracy}
\begin{figure*}
    \centering
    \includegraphics[width=460pt]{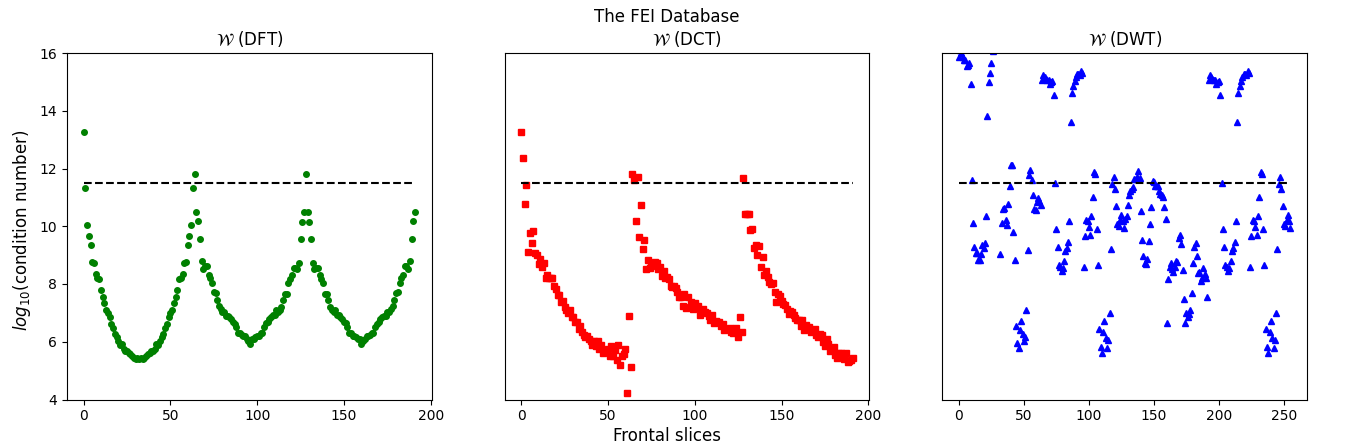}
\end{figure*}
\begin{figure*}
    \centering
    \includegraphics[width=460pt]{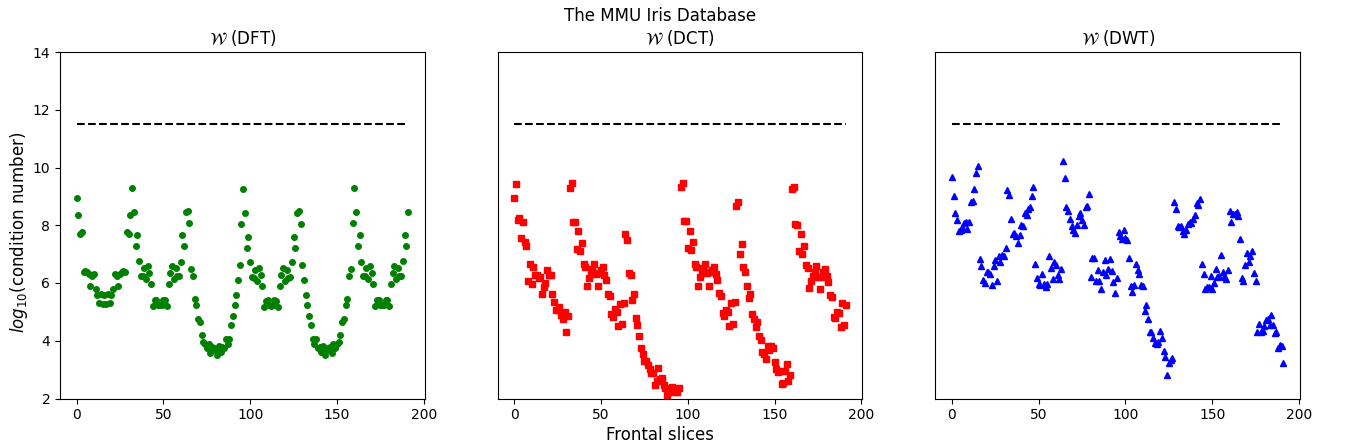}
\end{figure*}
\begin{figure*}
    \centering
    \includegraphics[width=460pt]{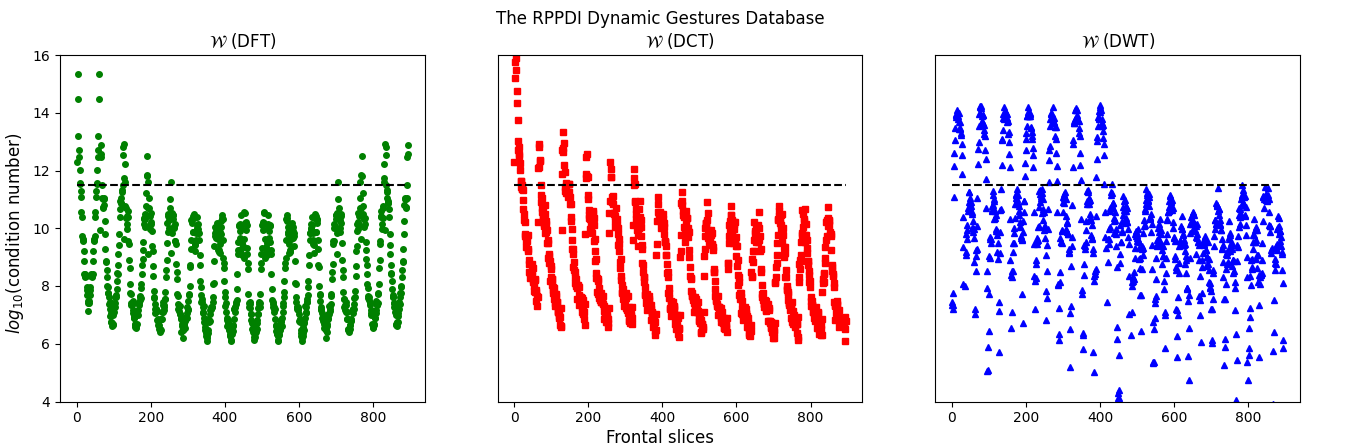}
\end{figure*}
\begin{figure*}
    \centering
    \includegraphics[width=460pt]{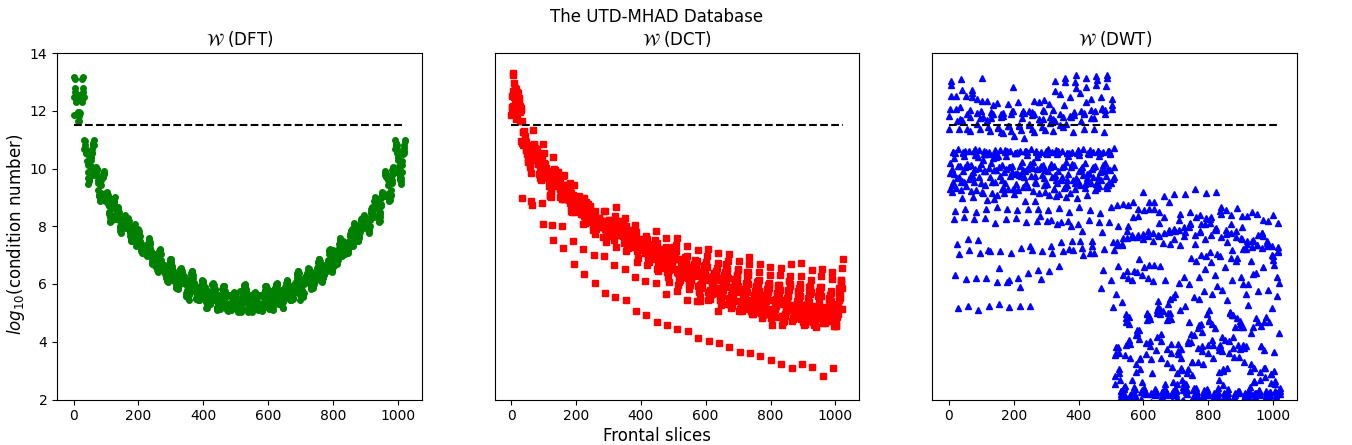}
    \caption{The condition numbers of the frontal slices of the within-class scatter tensors obtained by using the DFT, the DCT, and the DWT based tensor operators for the FEI, the MMU irish, the RPPDI dynamic gestures, and the UTD-MHAD data sets. The dashed lines show the threshold number $log_{10}(10^5)$.}
    \label{fig:condition}
\end{figure*}

\begin{table*}[!htbp]
\parbox{.49\textwidth}{
 \caption{Classification accuracy of each set for the FEI data set.}
 \label{tab:data_fei}
\centering
    \scalebox{0.85}{
         \begin{tabular}{c|c|c|c|c|c|c}
         \hline
          Methods& Set 1 & Set 2 &Set 3&Set 4&Set 5&mean$\pm$std\Tstrut\Bstrut \\
         \hline
         \hline
         HOMLDA-dft & ${92.86}$& $93.75$ & $91.43$&$93.93$&$91.25$&$92.64\pm1.13$\Tstrut\Bstrut \\
         RHOMLDA-dft & $94.29$ & $94.11$ & $92.14$&$93.21$&$91.07$&$92.96\pm1.21$\Tstrut\Bstrut\\
         \hline
         \hline
         HOMLDA-dct & $93.57$ & $94.29$ & $92.32$ & $93.75$ &$91.07$&$93.00\pm1.16$\Tstrut\Bstrut \\
         RHOMLDA-dct &$94.29$  & $94.11$ & $92.32$ &$92.68$ &$91.79$&$\mathbf{93.04\pm0.99}$\Tstrut\Bstrut \\
         \hline
         \hline
         HOMLDA-dwt & $90.89$ & $92.68$ & $89.46$ & $89.11$ &$88.57$&$90.14\pm1.48$\Tstrut\Bstrut \\
         RHOMLDA-dwt & $91.61$ & $92.50$ & $90.36$ & $90.18$ &$88.93$&$90.71\pm1.23$\Tstrut\Bstrut
         \\
         \hline
         \hline
         CMDA &$90.00$ & $91.79$  & $88.75$ & $92.68$&$86.79$&$90.00\pm2.11$\Tstrut\Bstrut\\
         \hline
         \hline
         DGTDA & $83.57$ & $84.82$ & $81.79$ & $84.11$ &$80.89$&$83.04\pm1.47$\Tstrut\Bstrut\\
         \hline
    \end{tabular}}
}
\hfill
\parbox{.49\textwidth}{
\caption{Classification accuracy of each set for the MMU iris data set.}
\label{tab:data_coil}
\centering
   \scalebox{0.85}{
         \begin{tabular}{c|c|c|c|c|c|c}
         \hline
          Methods& Set 1 & Set 2 &Set 3&Set 4&Set 5&mean$\pm$std\Tstrut\Bstrut \\
         \hline
         \hline
         HOMLDA-dft & ${88.89}$& $77.78$ & $86.67$&$84.44$&$80.00$&$83.56\pm4.12$\Tstrut\Bstrut \\
         RHOMLDA-dft & ${88.89}$& $77.78$ & $86.67$&$84.44$&$80.00$&$83.56\pm4.12$\Tstrut\Bstrut \\
         \hline
         \hline
         HOMLDA-dct & $82.22$ & $77.78$ & $88.89$ & $86.67$ &$86.67$&$84.44\pm3.98$\Tstrut\Bstrut \\
         RHOMLDA-dct & $82.22$ & $77.78$ & $88.89$ & $86.67$ &$86.67$&$\mathbf{84.44\pm3.98}$\Tstrut\Bstrut \\
         \hline
         \hline
         HOMLDA-dwt & $80.00$ & $66.67$ & $88.89$ & $88.89$ &$82.22$&$81.33\pm8.15$\Tstrut\Bstrut \\
         RHOMLDA-dwt& $80.00$ & $66.67$ & $88.89$ & $88.89$ &$82.22$&$81.33\pm8.15$\Tstrut\Bstrut \\
         \hline
         \hline
         CMDA &$86.67$ & $68.89$  & $84.44$ & $77.78$&$88.89$&$81.33\pm7.25$\Tstrut\Bstrut\\
         \hline
         \hline
         DGTDA & $77.78$ & $62.22$ & $86.67$ & $75.56$ &$82.22$&$76.89\pm8.27$\Tstrut\Bstrut\\
         \hline
    \end{tabular}}

}
\end{table*}

\begin{table*}[!htbp]
\parbox{.49\textwidth}{
\caption{Classification accuracy of each set for the RPPDI dynamic gestures data set.}
    \label{tab:data_cam}
\centering
    \scalebox{0.85}{
          \begin{tabular}{c|c|c|c|c|c|c}
         \hline
          Methods& Set 1 & Set 2 &Set 3&Set 4&Set 5&mean$\pm$std\Tstrut\Bstrut \\
         \hline
         \hline
         HOMLDA-dft & ${78.95}$& $76.32$ & $86.84$&$83.78$&$75.68$&$80.31\pm4.34$\Tstrut\Bstrut \\
         RHOMLDA-dft & $86.84$ & $94.74$ & $94.74$&$97.30$&$97.30$&${94.18\pm3.84}$\Tstrut\Bstrut\\
         \hline
         \hline
         HOMLDA-dct & $78.95$ & $78.95$ & $97.37$ & $81.08$ &$83.78$&$84.03\pm6.90$\Tstrut\Bstrut \\
         RHOMLDA-dct &$89.47$  & $94.74$ & $100.00$ &$97.30$ &$97.30$&$\mathbf{95.76\pm3.56}$\Tstrut\Bstrut \\
         \hline
         \hline
         HOMLDA-dwt & $81.58$ & $76.32$ & $89.47$ & $81.08$ &$86.49$&$82.99\pm4.57$\Tstrut\Bstrut \\
         RHOMLDA-dwt & $84.21$ & $81.58$ & $94.74$ & $83.78$ &$89.19$&$86.70\pm4.73$\Tstrut\Bstrut
         \\
         \hline
         \hline
         CMDA &$84.21$ & $78.95$  & $92.11$ & $89.19$&$91.89$&$87.27\pm5.04$\Tstrut\Bstrut\\
         \hline
         \hline
         DGTDA & $92.11$ & $73.68$ & $89.47$ & $78.38$ &$91.89$&$85.11\pm7.61$\Tstrut\Bstrut\\
         \hline
    \end{tabular}}
}
\hfill
\parbox{.49\textwidth}{
    \caption{Classification accuracy of each set for the UTD-MHAD data set.}
    \label{tab:data_ha}
\centering
   \scalebox{0.85}{
         \begin{tabular}{c|c|c|c|c|c|c}
         \hline
          Methods& Set 1 & Set 2 &Set 3&Set 4&Set 5&mean$\pm$std\Tstrut\Bstrut \\
         \hline
         \hline
         HOMLDA-dft & ${88.44}$& ${87.28}$ & ${89.60}$&${86.71}$&${85.47}$&$87.50\pm1.42$\Tstrut\Bstrut \\
         RHOMLDA-dft & ${95.95}$& ${94.80}$ & ${98.27}$&${98.84}$&${96.51}$&$96.87\pm1.50$\Tstrut\Bstrut \\
         \hline
         \hline
         HOMLDA-dct& ${91.91}$& ${87.28}$ & ${90.75}$&${91.33}$&${87.21}$&$89.70\pm2.03$\Tstrut\Bstrut \\
         RHOMLDA-dct & ${95.95}$& ${97.11}$ & ${98.27}$&${99.42}$&${96.51}$&$\mathbf{97.45\pm1.25}$\Tstrut\Bstrut \\
         \hline
         \hline
         HOMLDA-dwt  & ${91.91}$& ${92.49}$ & ${93.64}$&${91.91}$&${94.77}$&$92.94\pm1.11$\Tstrut\Bstrut \\
         RHOMLDA-dwt  & ${93.06}$& ${90.75}$ & ${94.80}$&${93.64}$&${95.93}$&${93.64\pm1.75}$\Tstrut\Bstrut \\
         \hline
         \hline
         CMDA  & ${91.33}$& ${91.91}$ & ${93.64}$&${89.02}$&${93.02}$&$91.78\pm1.60$\Tstrut\Bstrut \\
         \hline
         \hline
         DGTDA  & ${93.64}$& ${93.64}$ & ${95.95}$&${94.80}$&${96.51}$&$94.91\pm1.17$\Tstrut\Bstrut \\
         \hline
    \end{tabular}}

}
\end{table*}
We evaluate the classification accuracy of HOMLDA and RHOMLDA methods based on the DFT, the DCT and the DWT and the Tucker structure based MDA methods which are CMDA and DGTDA.
For each experiment, a $5$-fold cross-validation was applied and the classification accuracy was evaluated using the nearest neighbor search using the tensor Frobenious norm defined in \textbf{Definition~\ref{def:fro}}. 
TABLE~\ref{tab:data_fei},~\ref{tab:data_coil},~\ref{tab:data_cam}, and~\ref{tab:data_ha} show the classification accuracy of the FEI, the MMU iris, the RPPDI dynamic hand gesture, and the UTD-MHAD data sets respectively. The classification accuracy rates were evaluated for each set along with the mean classification accuracy $\pm$ the standard deviation in the classification accuracy rates across all $5$ sets. We compared our proposed methods with CMDA and DGTDA. As can be seen from the tables, the proposed RHOMLDA with the DCT (RHOMLDA-dct) approach
\footnote{Our source code is available in the GitHub repository: https://github.com/Cagri-Ozdemir/High-Order-Multilinear-Discriminant-Analysis}
outperforms the Tucker decomposition based approaches for each of the four data sets. 

As detailed in Section~\ref{subsec:RHOMLDA}, we determine the ill-condition frontal slices of the within-class scatter tensor and re-estimated all the frontal slices determined as ill-condition. The Fig.~\ref{fig:condition} shows the condition numbers of the frontal slices of the within-class scatter tensors obtained by using set-$1$ training samples for all the data sets used in the experiments. As can be seen from the plots, the within-class scatter tensor of the MMU iris data set does not need to be re-estimated, since all the frontal slices are determined as well-condition. That is why HOMLDA and RHOMLDA give the exact classification rates for the MMU iris data set. We can also observe from the TABLE~\ref{tab:data_cam} and TABLE~\ref{tab:data_ha} that once the DCT and the DFT based tensor operators are used for RHOMLDA, there is a significant incerase in the classification rates of  both the RPPDI dynamic gestures and the UTD-MHAD data sets. Thus, we can say that
distortion levels of the ill-conditioned within-class scatter tensors of HOMLDA-dft and HOMLDA-dct are significantly high for both the RPPDI dynamic gestures and the UTD-MHAD data sets.





\section{Conclusions and Future Work}
\label{sec:conc}
In this paper, we proposed a novel approach for multilinear discriminant analysis using tensor decomposition.
We defined a family of order-\textit{n} tensor operators and tensor eigendecomposition for any invertible linear transform. In particular, it was shown that using newly defined tensor operators, an order-\textit{n} tensor can be decomposed into the product of order-\textit{n} tensors, similar to a matrix eigenvalue decomposition.  Using these results, a new approach to multilinear  discriminant analysis was developed, referred to as HOMLDA. The proposed HOMLDA approach was formulated using three well-known linear transforms, namely the DFT, DCT, and DWT. The resulting framework, HOMLDA, was extended to RHOMLDA that provides a robust estimate of the inverse of the within-class scatter tensors.
An analysis was presented in the context of the classification of four different benchmark data sets, the FEI face data set, the MMU iris data set, the RPPDI dynamic gestures data set, and the UTD-MHAD data set. It was shown that our proposed RHOMLDA-dct method outperforms the state-of-the-art multilinear discriminant analysis methods based on Tucker decomposition structure. Future work will focus on extending the proposed work to a non-linear multilinear framework through a reproducing kernel Hilbert space by extending the well known ``kernel trick" to an order-\textit{n} tensor framework. Moreover, the authors would like to investigate the similarities between the subspaces of HOMLDA obtained via different invertible linear transformations via manifold projection algorithms and multidimensional scaling (MDS) for tensor objects.

\bibliographystyle{IEEEtran}
\bibliography{bibli.bib}
\begin{IEEEbiography}[{\includegraphics[width=25mm,height=32mm]{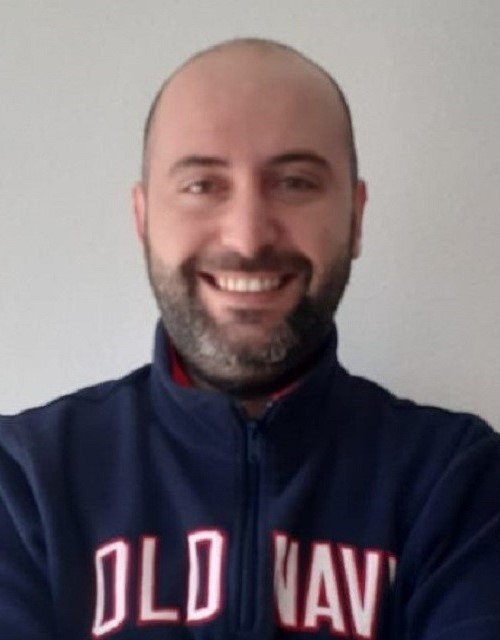}}]
{Cagri Ozdemir} (Student Member, IEEE) received the B.S degree from Dokuzeylul University in 2019. He is currently pursuing the Ph.D. degree in Computer Science and Engineering at South Dakota Mines. He joined South Dakota Mines, as a Research Assistant in 2020. His research interests include machine learning, higher-order data analysis, and signal processing.
\end{IEEEbiography}

\begin{IEEEbiography}[{\includegraphics[width=25mm,height=32mm]{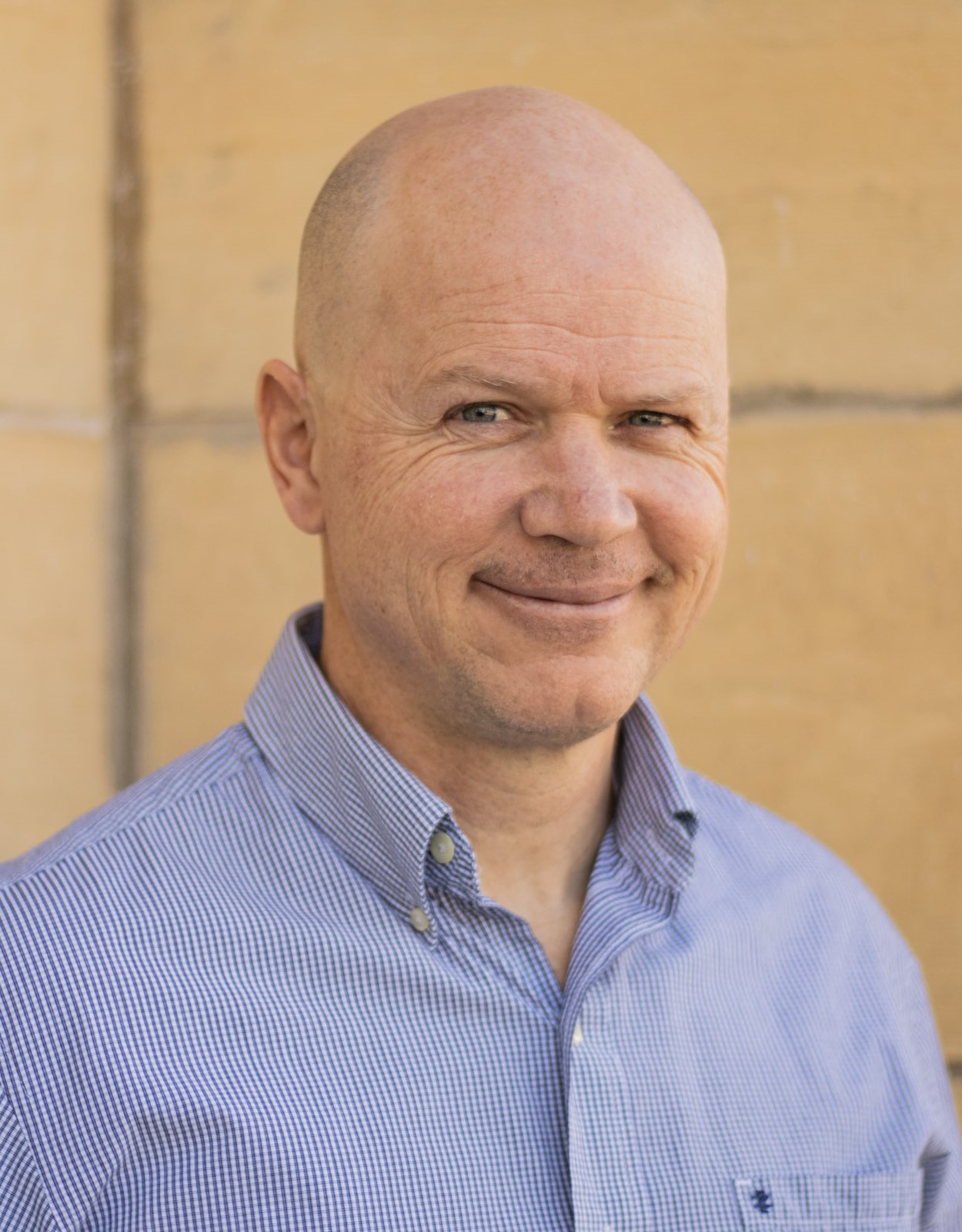}}]
{Randy C. Hoover} received the B.S degree in Electrical Engineering and M.S. degree in Measurement \& Control Engineering from Idaho State University in 2002 and 2005 respectively.  He received the Ph.D. in Electrical Engineering from Colorado State University in 2009.  He joined the faculty at South Dakota Mines in 2009 and spent nine years in the department of Electrical \& Computer Engineering before joining the Department of Computer Science and Engineering where he's currently an Associate Professor.  He was a National Science Foundation fellow from 2004 - 2005.  His research interests span machine learning, dimensionality reduction, multilinear systems, and multilinear subspace learning.
\end{IEEEbiography}

\begin{IEEEbiography}[{\includegraphics[width=25mm,height=32mm]{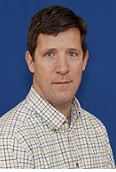}}]
{Kyle Caudle} received hs Ph.D. in Computational Statistics from George Mason University in 2006.  He joined the faculty at South Dakota Mines in 2011.  His research interests include streaming big data, density estimation and forecasting.  
\end{IEEEbiography}

\begin{IEEEbiography}[{\includegraphics[width=25mm,height=32mm]{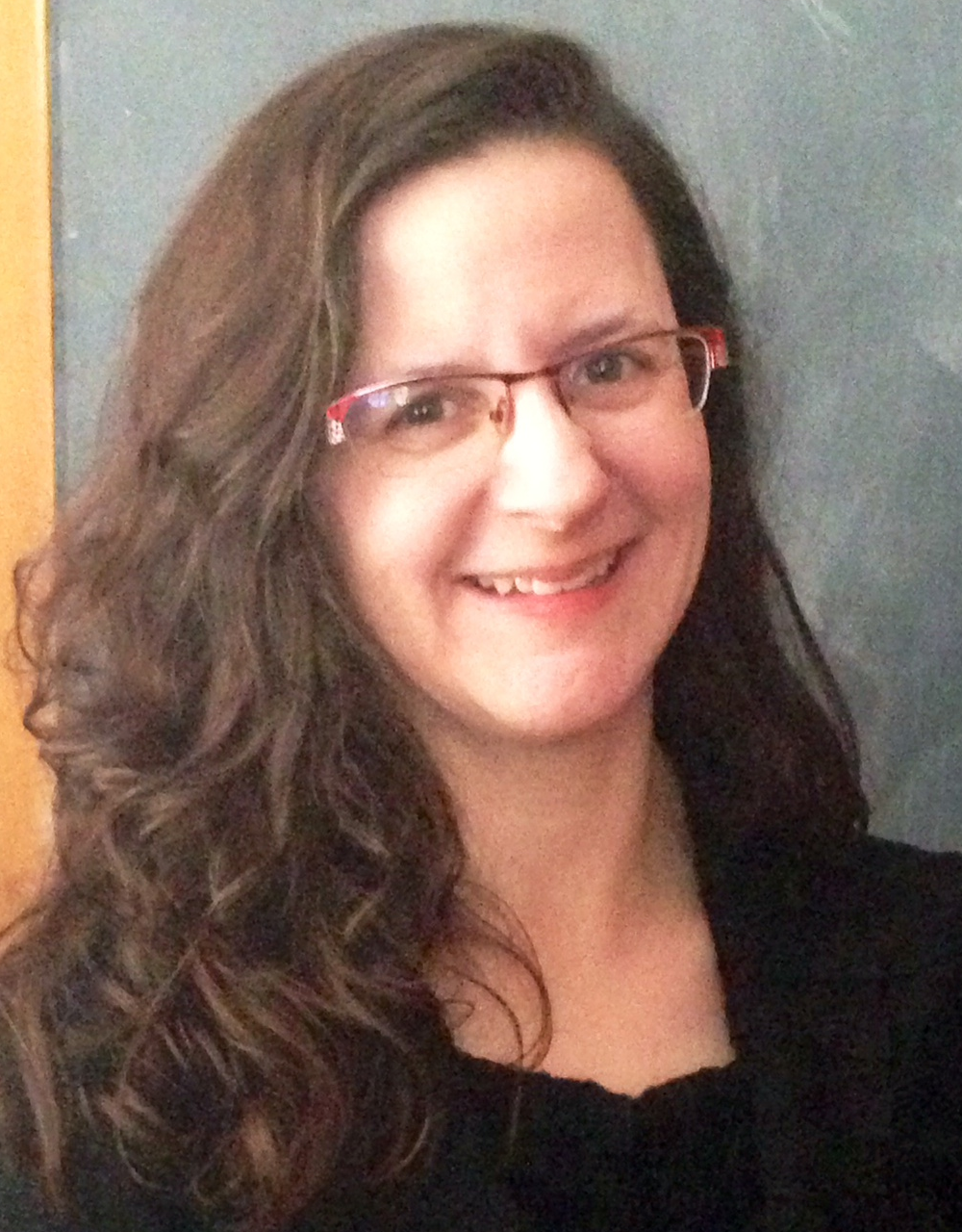}}]
{Karen Braman} received her Ph.D. in Mathematics from University of Kansas in 2003. She joined South Dakota Mines in 2004 and has been a Professor there since 2014. Her research interests include efficient solution of non-symmetric matrix eigenvalue problems and tensor decomposition theory and applications.
\end{IEEEbiography}

\end{document}